\documentclass[10pt,twocolumn,letterpaper]{article}
\usepackage{colortbl}
\usepackage[algorithms]{wacv}      

\usepackage{fancyhdr}
\definecolor{wacvblue}{rgb}{0.21,0.49,0.74}
\usepackage[pagebackref,breaklinks,colorlinks,allcolors=wacvblue]{hyperref}

\def\wacvPaperID{1011} 
\def\confName{WACV}
\def\confYear{2026}

\title{Feedback Alignment Meets Low-Rank Manifolds: A Structured Recipe for Local Learning}

\author{Arani Roy\\
Purdue University\\
{\tt\small roy173@purdue.edu}
\and
Marco P. Apolinario\\
Purdue University\\
{\tt\small mapolina@purdue.edu}
\and
Shristi Das Biswas\\
Purdue University\\
{\tt\small sdasbisw@purdue.edu}
\and
Kaushik Roy\\
Purdue University\\
{\tt\small kaushik@purdue.edu}
}

\begin{document}
\maketitle
\AtBeginDocument{\pagestyle{fancy}} 
\begin{abstract}
Training deep neural networks (DNNs) with backpropagation (BP) achieves state-of-the-art accuracy but requires global error propagation and full parameterization, leading to substantial memory and computational overhead. Direct Feedback Alignment (DFA) enables local, parallelizable updates with lower memory requirements but is limited by unstructured feedback and poor scalability in deeper architectures, specially convolutional neural networks. To address these limitations, we propose a structured local learning framework that operates directly on low-rank manifolds defined by the Singular Value Decomposition (SVD) of weight matrices. Each layer is trained in its decomposed form, with updates applied to the SVD components using a composite loss that integrates cross-entropy, subspace alignment, and orthogonality regularization. Feedback matrices are constructed to match the SVD structure, ensuring consistent alignment between forward and feedback pathways. Our method reduces the number of trainable parameters relative to the original DFA model, without relying on pruning or post hoc compression. Experiments on CIFAR-10, CIFAR-100, and ImageNet show that our method achieves accuracy comparable to that of BP. Ablation studies confirm the importance of each loss term in the low-rank setting. These results establish local learning on low-rank manifolds as a principled and scalable alternative to full-rank gradient-based training.
\end{abstract}
\section{Introduction}
The growing size and complexity of deep neural networks (DNN) have made their memory and computational demands a critical bottleneck, particularly in resource-constrained environments. Backpropagation (BP), today's dominant training method, amplifies these challenges due to its reliance on a global loss objective, the need to store intermediate activations, and the update-locking problem ~\cite{lillicrap2020backpropagation}, where the layer updates depend on the completion of both forward and backward passes. BP also suffers from the biologically implausible weight transport problem ~\cite{akrout2019deep,nokland2016direct}, which requires the transport of symmetric weights during the backward pass. These limitations highlight the need for alternative learning paradigms that offer parallelization, memory efficiency and hardware-compatibility.

\begin{figure}[t]
\begin{center}
\includegraphics[scale = 0.34]{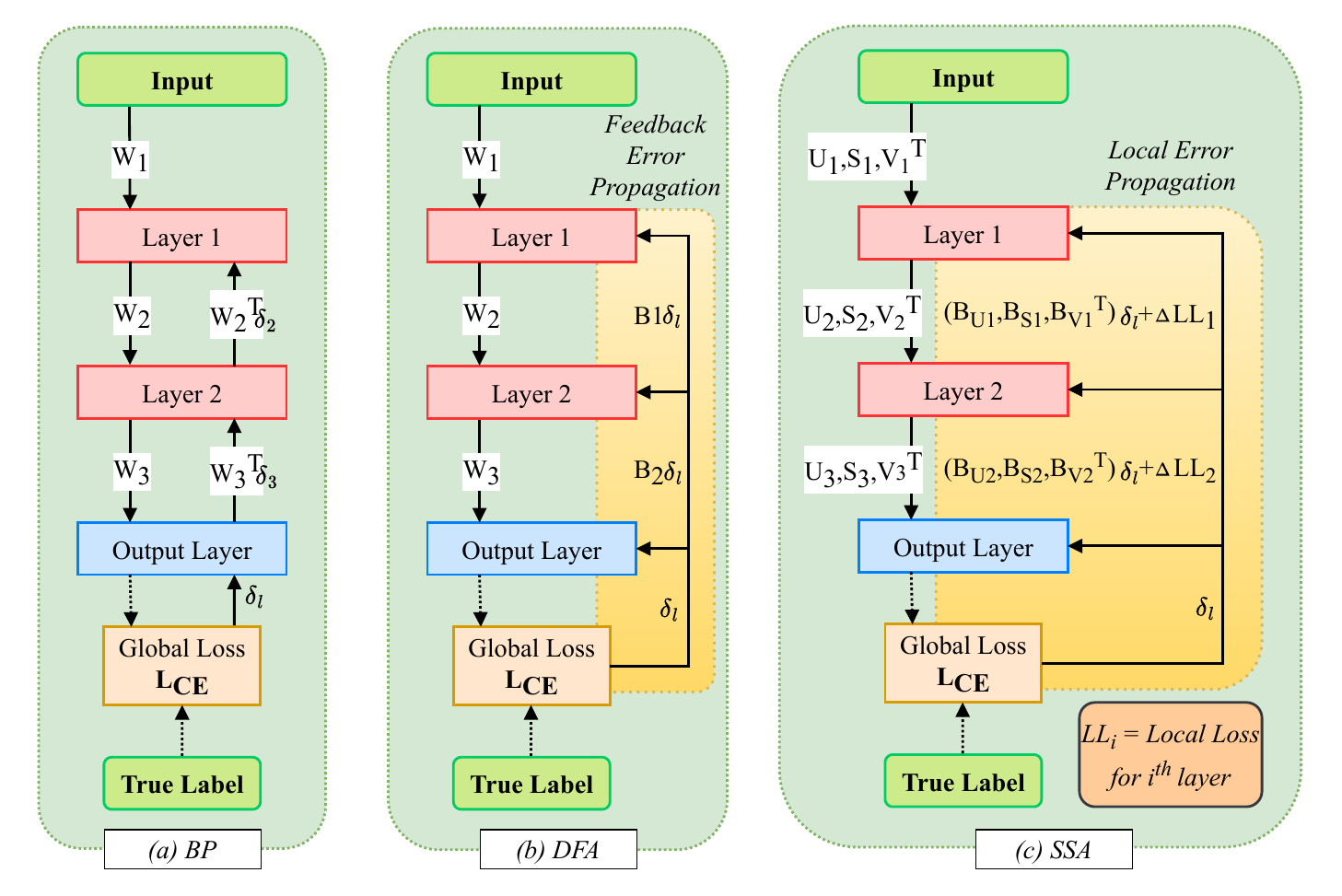}
\end{center}
\caption{A comparison of neural network training methods. Notations: $W$ = forward weights, $Layer$ = Layer Activations, $L_{CE}$ = cross-entropy loss, $B$ = random feedback weights, $\delta$ = gradients, $\delta_l$ = gradient of $L_{CE}$ = feedback error, $U,S,V^T$ = SVD components of forward weights, $B_U, B_S, B_{V^T}$ = SVD components of feedback weights, Local Loss (LL) has components: $L_{align}$ = Alignment loss, $L_{ortho}$ = Singular Vector Orthogonality regularizer. \textbf{(a) Backpropagation (BP):} Global gradient updates through each layer. \textbf{(b) Direct Feedback Alignment (DFA):} Local updates with random feedback \textbf{(c) SVD-space Alignment (SSA) (ours):} Decomposes weights into SVD components before training and aligns feedback on those decomposed components itself with local losses. Note: LL + $L_{CE}$ is the total local loss objective.}
\label{fig:compare}
\vspace{-1.5pt}
\end{figure}

Feedback Alignment (FA) ~\cite{lillicrap2016random} replaces the symmetric feedback weights with fixed random ones, showing that the forward weights align with them during training. Direct Feedback Alignment (DFA) ~\cite{nokland2016direct} extends this idea by removing both the weight symmetry requirement and update-locking constraint, transmitting output errors directly to each hidden layer. Unlike BP, DFA does not require the storage of intermediate activations or backward weight transport, significantly reducing memory and computational overhead. These characteristics make DFA particularly attractive for deployment on resource-constrained and neuromorphic hardware platforms~\cite{nokland2016direct,crafton2019direct,launay2020direct,neftci2017event}. Figure~\ref{fig:compare} summarizes the core differences in the update flow between BP and DFA. Despite its promise, DFA shows poor scalability to deeper architectures and convolutional networks. While some success has been observed in fully-connected layers~\cite{launay2020direct}, DFA fails on complex datasets such as CIFAR-100 and ImageNet without auxiliary strategies like transfer learning~\cite{crafton2019direct}. More recent improvements, such as the Direct Kolen–Pollack (DKP) algorithm~\cite{webster2021learning}, have improved alignment through learned symmetric feedback, but often at the cost of greater memory or training complexity. A key insight from~\cite{refinetti2021align} is that DFA’s effectiveness depends on the alignment between forward and feedback weights, an alignment that deteriorates with depth, especially in convolutional networks. Structural mismatches between fixed feedback paths and convolutional kernels further impede performance. Whether this alignment can be improved efficiently in low-rank subspaces remains an open question.
 \textbf{\textit{Could low-rank representations of weight and feedback spaces enable better alignment and extend DFA to convolutional layers? Could structured local updates make DFA scalable to deeper layers?}}  We hypothesize that modeling both forward and feedback signals within low-rank subspaces can increase alignment, improve inference efficiency, and enable scalable local learning in deep architectures, addressing key limitations of DFA.

We propose SVD-Space Alignment (SSA), a novel framework that combines the structural benefits of Singular Value Decomposition (SVD) with the efficiency of DFA. By operating directly on decomposed weight components ($U$, $S$, $V^T$), SSA enables structured updates within low-rank subspaces, aligning forward and feedback pathways to reduce randomness and improve convergence. A custom loss enforces task accuracy, alignment, and orthogonality. Furthermore, dynamic rank-scheduling scheme progressively compresses the model during training, reducing parameters below DFA levels without sacrificing performance. This synergy makes SSA a scalable and resource-efficient solution for training in constrained environments. In this work, we focus on demonstrating how low-rank constraints improve the scalability and stability of local learning.
Our local learning recipe offers the following key contributions:
\begin{itemize}
\item We propose SVD-Space Alignment (SSA), a novel local learning framework that combines Direct Feedback Alignment (DFA) with Singular Value Decomposition (SVD) to enable structured updates in low-rank subspaces, addressing DFA’s random feedback limitations.
\item We design a custom loss function combining feedback error, alignment loss, orthogonality regularization, and sparsity, to ensure efficient, convergent updates in the SVD-space.
\item We extend SSA to deep convolutional networks, demonstrating its ability to preserve spatial structure and hierarchical information while enabling scalable local learning.
\item Empirical results show that SSA achieves classification accuracy on par with BP, converges faster than DFA, and incurs lower inference cost than DFA on benchmarks such as CIFAR-100 and ImageNet.
\end{itemize}

\section{Background}
We consider a deep fully connected neural network with \( N \) layers. Each layer \( i \in \{1, \ldots, N\} \) has a weight matrix \( W_i \in \mathbb{R}^{d_i \times d_{i-1}} \), a pre-activation vector \( a_i \), and a post-activation output \( h_i \). Let \( f_i(\cdot) \) be the elementwise non-linearity at layer \( i \). The input to the network is \( h_0 = X \), and the final output is \( \hat{y} = h_N \).
The \textbf{forward pass} is defined recursively as:
\begin{equation}
\begin{aligned}
a_i &= W_i h_{i-1}, \\
h_i &= f_i(a_i), \quad \text{for } i = 1, \ldots, N.
\end{aligned}
\end{equation}

The task-specific loss \( \mathcal{L}(\hat{y}, y) \) quantifies the mismatch between the network's predictions \( \hat{y} \) and the true labels \( y \).

\textbf{Backward Pass with Backpropagation (BP):}
In standard BP, gradients are computed via recursive application of the chain rule. The error signal \( \delta a_i = \frac{\partial \mathcal{L}}{\partial a_i} \) is propagated backward from the top layer. Neglecting optimizer-specific details and biases, the gradient update for \( W_i \) is given by:
\begin{equation}
\delta W_i = -\frac{\partial \mathcal{L}}{\partial W_i}
= -\left[ \left( W_{i+1}^\top \delta a_{i+1} \right) \circ f_i'(a_i) \right] h_{i-1}^\top,
\label{eq:bp}
\end{equation}
where \( \circ \) denotes elementwise (Hadamard) product and \( f_i'(a_i) \) is the derivative of the activation function.
This recursive dependence on \( W_{i+1} \) breaks locality and requires symmetric weights, limiting parallelization and efficiency.

\textbf{Backward Pass with Direct Feedback Alignment (DFA):}
DFA replaces the non-local BP gradient signal \( W_{i+1}^\top \delta a_{i+1} \) with a random fixed projection of the final-layer error signal \( \delta a_N \). Each layer \( i \) uses a feedback matrix \( B_i \in \mathbb{R}^{d_i \times d_N} \), fixed at initialization, to compute its local update:
\begin{equation}
\delta W_i = -\left[ \left( B_i \delta a_N \right) \circ f_i'(a_i) \right] h_{i-1}^\top,
\label{eq:dfa}
\end{equation}
where \( \delta a_N = \frac{\partial \mathcal{L}}{\partial a_N} = \hat{y} - y \) for standard classification losses like mean squared error or cross-entropy. DFA removes upstream weight dependency, enabling local, parallel updates with lower memory and compute overhead.

\textbf{Alignment in DFA}
Although DFA uses fixed random feedback matrices, it has been empirically observed that the forward weights gradually adapt to align with the feedback pathways during training, such that \( W_i \approx \lambda B_i^\top \) for some scalar \( \lambda \). This emergent behavior, termed \emph{weight alignment}, improves the directional agreement between DFA’s pseudo-gradients and the true gradients computed via BP. More precisely, alignment refers to the similarity between the pseudo-gradient \( \delta a_i^{\text{DFA}} \) and the BP gradient \( \delta a_i^{\text{BP}} \). A common alignment metric is the cosine similarity between the DFA and BP weight gradients:
\begin{equation}
\cos\angle(\nabla_W \mathcal{L}_{\text{BP}}, \nabla_W \mathcal{L}_{\text{DFA}}) = \frac{\langle \nabla_W \mathcal{L}_{\text{BP}}, \nabla_W \mathcal{L}_{\text{DFA}} \rangle}{\|\nabla_W \mathcal{L}_{\text{BP}}\|_F \cdot \|\nabla_W \mathcal{L}_{\text{DFA}}\|_F}.
\end{equation}\label{eq:4}
Initially, due to random initialization, this alignment is weak. However, as training progresses, the forward weights rotate toward the subspace defined by the feedback matrices, leading to stronger alignment and more effective updates~\cite{refinetti2021align}. This phenomenon provides a key explanation for DFA’s empirical success despite bypassing gradient symmetry. Nevertheless, \cite{refinetti2021align} observe that this alignment tends to deteriorate in deeper networks, specially when structural mismatches exist between the fixed feedback paths and the convolutional or deep hierarchical structures of the forward weights. To address this, our method introduces structured forward and feedback paths and SVD-based loss terms that improve alignment and training stability, especially in convolutional and deep architectures.

\section{Related Works}
Our work builds on prior research in DFA, SVD-based representations in neural networks, and local layerwise training. This section reviews these foundations.

\textbf{DFA and its variants:}
DFA~\cite{nokland2016direct} offers a lightweight alternative to BP by replacing symmetric feedback with fixed random matrices, enabling local updates without backward weight transport. Despite its efficiency, DFA struggles with poor gradient alignment in deep and convolutional networks, limiting its scalability~\cite{sanfiz2021benchmarking, webster2021learning}. Enhancements such as auxiliary classifiers~\cite{nokland2019training} and partial integration into transformer models~\cite{launay2020direct} improve stability but retain BP in key components. To improve alignment, Uniform Sign-concordant Feedbacks (uSF) define $B_i = \text{sign}(W_i^\top)$, while Batchwise Random Magnitude Sign-concordant Feedbacks (brSF) set $B_i = |R_i| \cdot \text{sign}(W_i^\top)$ with resampled $|R_i|$. Our approach constrains our parameter space into low-rank manifolds, and adds custom local losses to the standard DFA errors, to improve alignment and scalability.

\textbf{SVD in Neural Networks:}
SVD has been widely used to compress neural networks by reducing the dimensionality of weight matrices, lowering parameter count and FLOPs with minimal accuracy loss~\cite{denil2013predicting}. Subsequent work~\cite{denton2014exploiting, yang2020learning} applied SVD to CNNs for inference acceleration, with~\cite{yang2020learning} introducing orthogonality regularization to stabilize training. However, these methods rely on backpropagation, which can disrupt the orthogonality of the decomposed weights. In contrast, SSA integrates SVD with DFA to preserve orthogonality and enables rank reduction during training, yielding compact models without losing accuracy.

\textbf{Local Layerwise Training:}
Local learning rules, which update layers independently of global gradients, are gaining attention for their scalability and efficiency. Methods like greedy layerwise training ~\cite{belilovsky2019greedy}, DRTP ~\cite{frenkel2021learning}, and AugLocal ~\cite{ma2024scaling}, reduce BP’s memory costs using local losses or auxiliary networks. Forward-only methods, such as Hinton's Forward-Forward (FF) ~\cite{hinton2022forward} and PEPITA ~\cite{dellaferrera2022error}, avoid backprop entirely but often fall short in accuracy, especially in deeper networks, and struggle with global alignment.
Our work focuses on improving local loss methods, as forward-only approaches struggle with classification accuracy in deeper networks. We enhance local layerwise training by applying SVD to layer weights and aligning representations with feedback matrices using DFA.

\section{Methodology}
We now present the details of SVD-Space Alignment (SSA), a local learning method that updates network weights in their SVD-decomposed form. Each layer’s orthogonal and singular components are optimized using a custom loss combining feedback error, alignment, and orthogonality regularization. Gradients are projected onto the tangent spaces of the Stiefel manifolds during training. The following subsections outline the structured forward and backward paths, loss formulation, SSA update rules, and dynamic rank scheduling. A theoretical perspective on our method is provided in Section 5.

\subsection{Embedding Structural Priors: Breaking Weights into SVD-Space }
Given a weight matrix $W_i \in \mathbb{R}^{m \times n}$ for layer $i$, we decompose it into its SVD form:
\begin{equation}
    W_i = U_i S_i V_i^T \label{eq:1}
\end{equation}
where $U_i \in \mathbb{R}^{m \times r}$ and $V_i \in \mathbb{R}^{n \times r}$ are orthogonal matrices, and $S_i \in \mathbb{R}^{r \times r}$ is a diagonal matrix of singular values, with rank $r \leq \min(m, n)$. Decomposing weights prior to training allows updates to be applied directly in the SVD-space, preserving the orthogonality of $U_i$ and $V_i^T$ throughout the training process. This approach avoids the computational overhead associated with performing SVD at every epoch, while facilitating efficient rank reduction during training.

\textbf{SVD-Space for Convolutional Layers:}  
To adapt the SVD-space approach for convolutional layers, we employ a spatial decomposition strategy inspired by ~\cite{yang2020learning}. A convolutional kernel $K \in \mathbb{R}^{N \times C \times H \times W}$, with $N$ as the number of filters, $C$ as the number of input channels, and $H \times W$ as the spatial dimensions, is first reshaped into a 2D matrix $K^\prime \in \mathbb{R}^{NW \times CH}$. $K^\prime$ is then decomposed using SVD: $K^\prime = U \Sigma V^T$ where $U \in \mathbb{R}^{NW \times r}$ and $V \in \mathbb{R}^{CH \times r}$ are unitary matrices, and $\Sigma \in \mathbb{R}^{r \times r}$ is a diagonal matrix of singular values, with rank $r = \min(NW, CH)$. The decomposed components are reshaped into convolutional layers as follows: $U \sqrt{\Sigma}$ is reshaped into a convolutional kernel $K_1 \in \mathbb{R}^{r \times C \times H \times 1}$, $\sqrt{\Sigma} V^T$ is reshaped into a kernel $K_2 \in \mathbb{R}^{N \times r \times 1 \times W}$.
This decomposition effectively factorizes the original convolution into two consecutive operations. The first convolution, with $K_1$, reduces the input channels and captures spatial information across the height dimension, while the second convolution, with $K_2$, restores the original output dimensionality and captures spatial information across the width dimension. This hierarchical decomposition effectively preserves spatial and hierarchical features. We retain the decomposed kernels for inference. For SSA, we form our feedback matrices to be of the same dimensions as our convolutional kernels.

\subsection{Local Loss Objective Design}
The local loss function at layer $i$ is designed to ensure efficient and structured learning directly in the SVD-space. It consists of three key components:
\begin{equation}
    LL_i(\theta_i) = \alpha L_{\text{CE}}(\theta_i) + \beta L_{\text{align}}(\theta_i) + \gamma L_{\text{ortho}}(\theta_i)
\end{equation}
where $\theta_i = (U_i, S_i, V_i^T)$, and each term contributes uniquely to the training process. We provide the gradients of this loss to the layer as shown in Fig. 1 (c).

\textbf{Cross-Entropy Loss ($L_{\text{CE}}$):} This term ensures the model learns the primary task like classification. In the DFA framework, the propagated feedback error is derived directly from the gradient of the model cross-entropy loss:
\begin{equation}
    \Delta L_{\text{CE}} = y_{\text{predict}} - y_{\text{label}}
\end{equation}
Here, $\Delta L_{\text{CE}} \ i.e.\  \delta_l$ guides the local updates in DFA and aligns predictions with the target outputs.

\textbf{Alignment Loss ($L_{\text{align}}$):} This term ensures alignment between the forward SVD components $(U_i, S_i, V_i^T)$ and the feedback matrices $(B_{U_i}, B_{S_i}, B_{V_i^T})$ used for DFA updates, reducing mismatch in SVD-space. The loss is defined as:
\begin{equation}
    L_{\text{align}}(\theta_i) = \|U_i - B_{U_i}\|_F^2 + \|S_i - B_{S_i}\|_F^2 + \|V_i^T - B_{V_i^T}\|_F^2
\end{equation}

\textbf{Singular Vector Orthogonality Regularizer ($L_{\text{ortho}}$):} To ensure the numerical stability of the SVD decomposition and a well-structured representation during training, this regularizer promotes orthogonality in the singular vectors $U_i$ and $V_i^T$. It is formulated as:
\begin{equation}
    L_{\text{ortho}}(\theta_i) = \|U_i^T U_i - I\|_F^2 + \|V_i^T V_i - I\|_F^2
\end{equation}

\subsection{SSA Updates in SVD-Space}
In SSA, DFA is adapted to operate directly in the SVD-space of the weight matrices \eqref{eq:1}. 
For each layer $i$, we first sample a fixed random feedback matrix 
$B_i \in \mathbb{R}^{d_i \times d_N}$ from a Gaussian distribution 
$\mathcal{N}(0, \sigma^2)$, followed by column-wise normalization to unit $\ell_2$ norm. 
To impose structure, we compute a truncated randomized SVD of $B_i$: 
$B_i \;\approx\; B_{U_i} \, B_{S_i} \, B_{V_i}^\top$,
where the retained rank $r$ is chosen to match the forward decomposition. These structured factors $(B_{U_i}, B_{S_i}, B_{V_i})$ serve as explicit alignment targets in the local loss objective.
The update rules for each component are as follows:
\begin{equation}
\begin{aligned}
 & U_i^{(t+1)} = U_i^{(t)} - \eta \nabla_{U_i} LL_i(U_i, S_i, V_i^T)
\\ & S_i^{(t+1)} = S_i^{(t)} - \eta \nabla_{S_i} LL_i(U_i, S_i, V_i^T)
\\ & V_i^{T(t+1)} = V_i^{T(t)} - \eta \nabla_{V_i^T} LL_i(U_i, S_i, V_i^T)
\end{aligned}
\end{equation}
where $\eta$ is the learning rate, and $LL_i(U_i, S_i, V_i^T)$ is the local loss for layer $i$. These updates are fully local and parallelizable, requiring no global error propagation.
The gradients for each SVD component are derived from the local loss $LL_i$ and are detailed below.

\textbf{Singular Values \(S_i\):}
\begin{equation}
\nabla_{S_i} LL_i = U_i^\top \nabla_{W_i} L_{\text{CE}} V_i + 2\beta (S_i - B_{S_i})
\end{equation}

\textbf{Left Singular Vectors \(U_i\):}
\begin{equation}
\begin{aligned}
\nabla_{U_i} LL_i = (I - U_i U_i^\top) \big[ 
& \nabla_{W_i} L_{\text{CE}} V_i S_i^{-1} \\
& + 2\beta (U_i - B_{U_i}) \\
& + 4\gamma U_i (U_i^\top U_i - I) 
\big]
\end{aligned}
\end{equation}

\textbf{Right Singular Vectors \(V_i^\top\):}
\begin{equation}
\begin{aligned}
\nabla_{V_i^\top} LL_i = (I - V_i^\top V_i) \big[
& \nabla_{W_i} L_{\text{CE}}^\top U_i S_i^{-1} \\
& + 2\beta (V_i^\top - B_{V_i^\top}) \\
& + 4\gamma V_i^\top (V_i V_i^\top - I)
\big]
\end{aligned}
\end{equation}

The projection terms \( (I - U_i U_i^\top) \) and \( (I - V_i^\top V_i) \) ensure updates remain on the Stiefel manifold (details in 5.2), preserving orthogonality during optimization.

\subsection{Dynamic Rank Reduction Strategy}
SVD-based parameterization increases the model size from \( mn \) to approximately \( r(m + n) \), effectively doubling the parameter count when \( r = \min(m, n) \). To ensure SSA remains compact and comparable to the original model, we employ a hybrid rank reduction strategy, where the effective rank \( r \) is constrained to satisfy:
$r < \frac{mn}{m+n}$
Training begins with full rank \( r_0 \). Over the first 30\% of epochs, we apply an \textbf{epoch-based schedule} that reduces \( r \) to \( 0.7r_0 \). This is guided by the Hoyer regularizer, applied every 10 epochs to promote sparsity in singular values: $\mathcal{L}_{\text{Hoyer}}(S_i) = \frac{\|S_i\|_1}{\|S_i\|_2}$.
In later epochs, we adopt a \textbf{threshold-based schedule} that retains the top singular values accounting for 95\% of the spectral energy, preventing over-pruning near convergence. While the epoch-based schedule enables rapid early compression, reducing the rank below 30\% of \( r_0 \) often destabilizes training. Our hybrid strategy balances early aggressive pruning with stable, energy-aware refinement. Notably, SSA trains predominantly and always infers at reduced rank, demonstrating its efficiency in both convergence and deployment.

\section{Theoretical Foundations and Alignment Properties in SSA}
In this section, we examine the theoretical foundations and alignment characteristics of our method. We begin by analyzing the properties of the local loss, followed by a detailed explanation of the update procedures on the Stiefel manifold. We then discuss the relationship between SSA and BP gradients, providing evidence that SSA achieves stronger gradient alignment than DFA. Finally, we study the dynamics of the matrix alignment angle across training epochs and network depth, illustrating how SSA maintains more stable alignment in deeper architectures.

\subsection{SSA Local Loss Properties}
Convexity and smoothness of the loss function are essential for stable and efficient optimization. The cross-entropy loss \( L_{\text{CE}} \) is convex and \( L \)-smooth with Lipschitz constant \( L = 0.5 \), while the alignment loss \( L_{\text{align}} \), composed of squared Frobenius norms, is also convex and \( 2 \)-smooth~\cite{shalev2014understanding,boyd2004convex}. In contrast, the orthogonality regularizer \( L_{\text{ortho}} \), which involves quartic terms such as \( \|U^\top U - I\|_F^2 \), is non-convex. To address this, we constrain the updates of \( U \) and \( V \) to lie on the Stiefel manifold~\cite{schotthofer2022low}, projecting gradients onto its tangent space to preserve orthogonality and improve optimization stability. While the overall objective is non-convex due to these constraints, the dominance of convex and smooth terms in the local layerwise losses results in empirically stable and efficient convergence. A detailed analysis of convexity and smoothness properties is provided in Supplementary Section 1. Although a global convergence guarantee is difficult in this setting, we observe that maintaining local Lipschitz continuity and smoothness strongly correlates with stable training and improved final performance in practice.

\subsection{Orthogonality-Preserving Updates}
The Stiefel manifold \( \mathbb{V}_{r}(\mathbb{R}^n) \) comprises all \( n \times r \) matrices with orthonormal columns, i.e., \( U^\top U = I \). To preserve this structure in the singular matrices \( U_i \) and \( V_i \), SSA employs a two-fold strategy: (1) projection-based gradient updates and (2) soft orthogonality regularization.
We project the gradients onto the tangent space of the Stiefel manifold as:
\[
\nabla_{U_i} \mathcal{L} \leftarrow (I - U_i U_i^\top) \nabla_{U_i} \mathcal{L}, \quad
\nabla_{V_i^T} \mathcal{L} \leftarrow (I - V_i^T V_i) \nabla_{V_i^T} \mathcal{L}.
\]
This lightweight operation avoids costly retractions (QR or Cayley transforms), while preserving local orthogonality constraints. It aligns with recent findings~\cite{ablin2022fast, schotthofer2022low} that such projection-only methods can match full Riemannian optimizers in convergence and constraint satisfaction. To mitigate long-term drift due to stochastic updates or numerical imprecision, we additionally include a soft penalty \( \|U_i^\top U_i - I\|_F^2 \) and \( \|V_i V_i^\top - I\|_F^2 \) in the loss function. This complementary design enables stable, structure-aware optimization in the SVD space throughout training.

\subsection{Alignment Between SSA and BP Gradients}
As per ~\cite{nokland2016direct} and ~\cite{bordelon2022influence}, alignment refers to the degree of similarity between BP gradient and FA techniques based pseudo-gradients. High alignment (alignment angle $>0^{\circ}$) suggests a better approximation of the true direction of descent of the gradient. To demonstrate that SSA gradients align with BP gradients, we analyze the cosine similarity between the true gradient $\nabla_W L_{\text{true}}$ and the SSA gradient $\nabla_W L_{\text{SSA}}$. Expanding $\nabla_W L_{\text{SSA}}$:
\begin{equation}
\nabla_W L_{\text{SSA}} = \nabla_W L_{\text{CE}} + 2\beta \nabla_W L_{\text{align}} + 4\gamma \nabla_W L_{\text{ortho}}
\end{equation}
We analyze the contributions of each term:

\textbf{Contribution from $L_{\text{CE}}$:} 
The primary descent direction is dictated by $\nabla_W L_{\text{CE}}$ (original DFA feedback error), which matches $\nabla_W L_{\text{true}}$ as shown in DFA prior works ~\cite{nokland2016direct}. 

\textbf{Contribution from $L_{\text{align}}$:} 
The gradient of the alignment loss is:
\begin{equation}
\nabla_W L_{\text{align}} = 2(W - B)
\end{equation}
Since $B$ approximates $W^T$ for the backward gradient pathway, this term complements $\nabla_W L_{\text{CE}}$, ensuring positive alignment. This assumes that $W$ and $B$ are reasonably aligned over time, and the difference between them is small, given proper initialization of $B$.

\textbf{Contribution from $L_{\text{ortho}}$:}
The orthogonality regularizer ensures stability by constraining the singular vectors to the Stiefel manifold. The gradient of $L_{\text{ortho}}$ is tangent to this manifold, so:
\begin{equation}
\langle \nabla_W L_{\text{true}}, \nabla_W L_{\text{ortho}} \rangle \approx 0
\end{equation}
Also, the orthogonality-preserving projections contribute to stability without altering descent direction.

The inner product between $\nabla_W L_{\text{true}}$ and $\nabla_W L_{\text{SSA}}$ is:
\begin{equation}
\langle \nabla_W L_{\text{true}}, \nabla_W L_{\text{SSA}} \rangle = \|\nabla_W L_{\text{true}}\|^2 + 2\beta \langle \nabla_W L_{\text{true}}, W - B \rangle
\end{equation}
This is guaranteed to be positive because $\|\nabla_W L_{\text{true}}\|^2 > 0$ and $\langle \nabla_W L_{\text{true}}, W - B \rangle > 0$.

Thus:
\begin{equation}
\cos(\nabla_W L_{\text{true}}, \nabla_W L_{\text{SSA}}) > 0
\end{equation}
ensuring the angle between the gradients is $\leq 90^\circ$. This alignment guarantees that SSA updates descend along a similar direction to the true gradient of BP.

\subsection{Gradient Alignment Analysis}
\begin{figure}[t]
    \centering
    \includegraphics[width=0.45\textwidth]{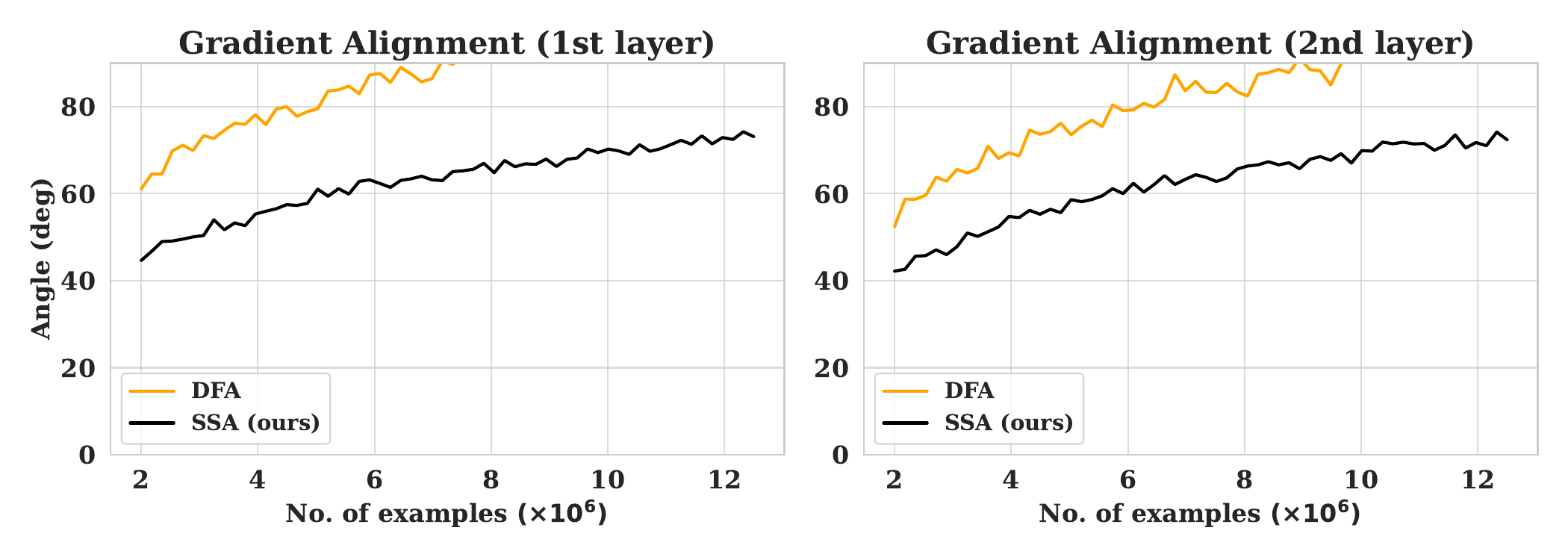}
    \caption{Gradient alignment (angle in degrees) between SSA/DFA and BP gradients across epochs on a 3-layer MLP. Lower angles indicate better alignment.}
    \label{fig:gradient_alignment}
\end{figure}

To assess whether SSA produces updates more aligned with the true supervised gradient than DFA, we compute the alignment angle (in degrees) between each method's update and the corresponding backpropagation (BP) gradient. We run a 3-layer MLP on CIFAR-10, and log alignment angles for each layer for different sample sizes. We compare against DFA to quantify how incorporating structure and Stiefel-regularized updates in SSA affects alignment. As shown in Figure~\ref{fig:gradient_alignment}, SSA consistently achieves lower alignment angles across layers compared to DFA, indicating that SSA updates are better aligned with the true BP direction.

\subsection{Layerwise Alignment Analysis over Time}
\begin{figure}[t]
    \centering
    \includegraphics[width=0.5\textwidth]{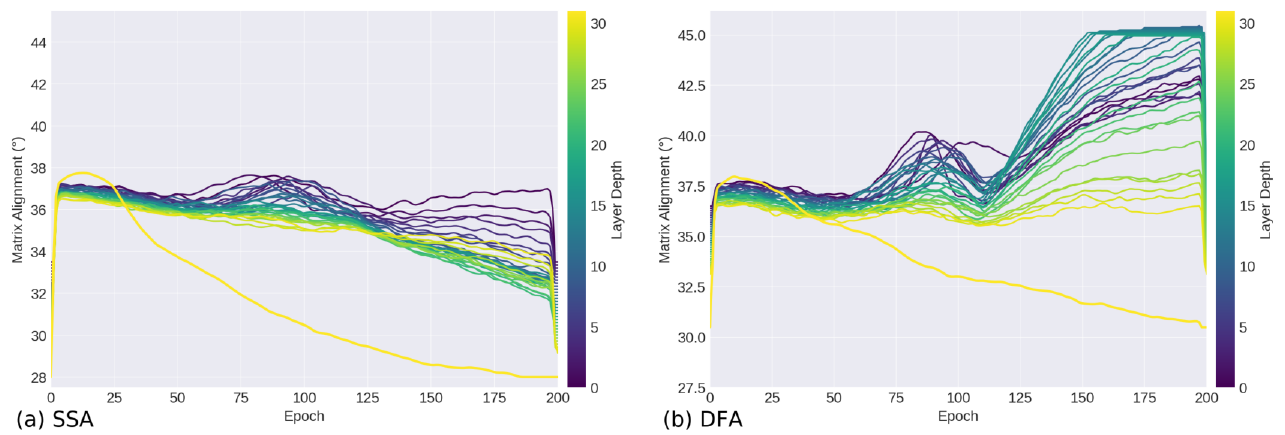}
    \caption{Matrix alignment angles across ResNet-32 layers over training epochs, between (a) SSA (b) DFA, and BP}
    \label{fig:matrix_alignment}
\end{figure}
Additionally, we evaluate the \emph{matrix alignment angle}, defined as the geometric angle between the column spaces of $W^\top$ and $B$, with smaller angles indicating stronger alignment. Tracking this metric across training epochs and network depth in figure ~\ref{fig:matrix_alignment} reveals how feedback alignment evolves during learning. As expected, the \emph{last layer closest to the output} achieves the best alignment, since error information is directly available at the output. In deeper layers, the alignment varies more gradually across epochs, reflecting the increasing complexity of propagating structured signals through the network. Crucially, SSA outperforms DFA by maintaining lower alignment angles in deeper layers, yielding greater stability and reduced error. This highlights SSA as a principled and more reliable alternative to DFA in deep networks.

\section{Experimental Setup}
We evaluate the performance of our framework by outlining the datasets, architectures, baselines, and evaluation metrics for computer vision tasks.

\textbf{Datasets.} We conduct experiments on CIFAR-10, CIFAR-100~~\cite{krizhevsky2009learning}, and ImageNet (ILSVRC-2012)~~\cite{krizhevsky2017imagenet}, using CIFAR for small-scale benchmarks and ImageNet for large-scale evaluation.

\textbf{Network Architectures.} SSA is evaluated on SmallConv (conv96-pool-conv192-pool-conv512-pool-fc1024), VGG-13~~\cite{simonyan2014very}, and ResNet-32~~\cite{he2016deep}, covering diverse depths and sizes to assess flexibility and scalability.

\textbf{Baselines.} We compare SSA against BP, DFA~~\cite{nokland2016direct}, SVD-BP (low-rank BP without feedback alignment), PredSim~~\cite{nokland2019training}, AugLocal~~\cite{ma2024scaling}, DRTP~~\cite{frenkel2021learning}, and PEPITA~~\cite{dellaferrera2022error}, evaluating accuracy, memory, computational cost, and training stability.

\textbf{Evaluation Metrics.} We measure the following.
\textbf{Classification Accuracy:} Top-1 and top-5 accuracy for generalization.
\textbf{Memory Usage:} Training memory footprint for efficiency.
\textbf{Computational Cost:} FLOPs per training iteration to quantify efficiency gains.

\textbf{Training Details.} Experiments are conducted on NVIDIA A40 GPUs with 48 GB memory, implemented in PyTorch. Training uses the Adam optimizer~~\cite{kingma2014adam} with a learning rate between $1\text{e}{-4}$ and $5\text{e}{-4}$. Data augmentation includes random cropping and horizontal flipping. Batch sizes are 128 for CIFAR and 256 for ImageNet. SSA starts with full-rank weight matrices and progressively reduces rank during training as described in Section 4.4.

\section{Results and Analysis}
In this section, we present comprehensive experimental results to evaluate the effectiveness of the proposed SSA method. We benchmark SSA against baseline methods, including standard DFA and its recent variants, across multiple metrics. An ablation study is conducted to assess the contribution of individual loss components. Beyond accuracy and convergence behavior, we also quantify the memory and computational gains enabled by SSA’s structured learning and low-rank optimization.

\subsection{Classification Accuracy}
We evaluate the classification accuracy of SSA on CIFAR-10 and ImageNet and compare it against a range of baselines. Table~\ref{tab:smallconv_comparison} reports results for SmallConv, including BP, SVD-BP, and several local or forward-only methods. Here, SVD-BP trains the SVD components using backpropagation, DRTP projects the targets (rather than the global error) to local layers, and PEPITA trains layers via two forward passes. The Forward-Forward method is excluded as it does not extend to convolutional networks. SSA outperforms the other local and forward-only methods while reaching accuracy comparable to BP.

\begin{table}[htbp]
\centering
\Large
\rowcolors{1}{yellow!15}{yellow!15}
\resizebox{\linewidth}{!}{
\begin{tabular}{l|c|c}
\hline
\textbf{Method} & \textbf{CIFAR-10 (mean $\pm$ std)} & \textbf{CIFAR-100 (mean $\pm$ std)} \\ \hline
 BP          & $87.57 \pm 0.14$ & $62.25 \pm 0.21$ \\  
SVD-BP ~\cite{yang2020learning}      & $87.30 \pm 0.18$ & $61.64 \pm 0.19$ \\  
DFA ~\cite{akrout2019deep}        & $73.10 \pm 0.53$ & $44.93 \pm 0.51$ \\  
DRTP ~\cite{frenkel2021learning}       & $68.96 \pm 0.80$ & NA    \\  
PEPITA ~\cite{dellaferrera2022error}     & $56.34 \pm 1.24$ & $27.56 \pm 0.67$ \\  
SSA (ours)  & $86.21 \pm 0.10$ & $60.72 \pm 0.15$ \\ \hline
\end{tabular}
}
\caption{Comparison of classification accuracy (mean $\pm$ standard deviation) over 5 independent runs with random inits for CIFAR-10 and CIFAR-100 datasets, with SmallConv network.}
\label{tab:smallconv_comparison}
\end{table}

\begin{table*}[htbp]
\centering
\rowcolors{1}{blue!5}{blue!5}
\resizebox{\textwidth}{!}{%
\begin{tabular}{l|l|c|c|c}
\hline
\textbf{Network} & \textbf{Method} & \textbf{CIFAR10 (Top-1, mean $\pm$ std)} & \textbf{ImageNet (Top-1, mean $\pm$ std)} & \textbf{ImageNet (Top-5, mean $\pm$ std)} \\ \hline
\textbf{VGG-13}  & BP                        & $93.75 \pm 0.12$ & $71.59 \pm 0.25$ & $90.39 \pm 0.14$ \\  
                 & SVD-BP~\cite{yang2020learning} & $92.80 \pm 0.15$ & $71.37 \pm 0.20$ & $90.20 \pm 0.18$ \\  
                 & PredSim~\cite{nokland2019training} & $86.49 \pm 0.53$ & NA              & NA              \\  
                 & AugLocal~\cite{ma2024scaling}    & $93.72 \pm 0.10$ & $70.93 \pm 0.22$ & $90.16 \pm 0.16$ \\  
                 & SSA (ours)                 & $92.70 \pm 0.13$ & $69.68 \pm 0.24$ & $88.84 \pm 0.15$ \\ \hline
\textbf{ResNet-32}  & BP                        & $92.14 \pm 0.11$ & $74.28 \pm 0.30$ & $91.76 \pm 0.12$ \\  
                    & SVD-BP~\cite{yang2020learning} & $91.77 \pm 0.14$ & $72.91 \pm 0.27$ & $89.27 \pm 0.20$ \\  
                    & PredSim~\cite{nokland2019training} & $79.31 \pm 0.45$ & NA              & NA              \\  
                    & AugLocal~\cite{ma2024scaling}    & $93.47 \pm 0.12$ & $73.95 \pm 0.25$ & $91.70 \pm 0.17$ \\  
                    & SSA (ours)                 & $88.02 \pm 0.18$ & $69.38 \pm 0.23$ & $87.72 \pm 0.19$ \\ \hline
\end{tabular}%
}
\caption{Comparison of classification accuracy (mean $\pm$ standard deviation) over 5 independent runs for CIFAR-10 and ImageNet datasets.}
\label{tab:vgg_resnet_comparison}
\end{table*}
Table~\ref{tab:vgg_resnet_comparison} extends the comparison to larger networks (VGG-13 and ResNet-32) on CIFAR-10 and ImageNet. On VGG-13, SSA matches BP within 0.2\% on CIFAR-10 and remains competitive on ImageNet, showing that SSA scales effectively where DFA and other local methods from Table~\ref{tab:smallconv_comparison} collapse. We include DFA here to highlight this scalability gap, while omitting other baselines that fail entirely at this scale. On ResNet-32, SSA exhibits a larger gap to BP, reflecting a common limitation of local learning: feedback signals weaken as they traverse many layers without gradient information, and residual connections further distort them. Nevertheless, SSA still achieves substantially higher accuracy than DFA, demonstrating that structure-aware learning mitigates, though does not completely resolve, these challenges. For completeness, we note that PredSim combines prediction and similarity matching losses but does not report ImageNet results, while AugLocal attains strong accuracy by embedding later-layer information into earlier layers at the cost of substantial auxiliary overhead. In contrast, SSA achieves comparable accuracy without such additions, underscoring its efficiency and scalability.

\subsection{Ablation Study}
We perform an ablation study to quantify the contribution of each component in the composite loss function to the overall performance of SSA. Table~\ref{table:ablation} reports the impact of removing each term: Cross-Entropy Loss, Alignment Loss, and the Orthogonality Regularizer, on classification accuracy and computational efficiency. Additional results, including a sensitivity analysis of loss coefficients \( \alpha, \beta, \gamma \), are provided in the supplementary material.

\begin{table}[htbp]
\centering
\small
\caption{Ablation study on CIFAR-10 showing the impact of each component in the composite loss function.}
\label{table:ablation}
\begin{tabular}{l|c|c}
\hline
\textbf{Component Removed}      & \textbf{Accuracy (\%)} & \textbf{FLOPs (B)} \\ \hline
Full SSA (All Components)       & 92.70                   & 0.14               \\
No Cross-Entropy Loss           & 27.05                   & 0.133              \\
No Alignment Loss               & 83.12                   & 0.119              \\
No Orthogonality Regularizer    & 85.44                   & 0.112              \\ \hline
\end{tabular}
\end{table}

From Table \ref{table:ablation}, removing the Cross-Entropy Loss results in a significant accuracy drop, as expected. Alignment loss attempts to reduce the loss between the subspaces in forward and feedback weights. Alignment loss aids gradient direction preservation, and therefore, removing these components also decreases accuracy, but less severely than Cross-Entropy Loss. Orthogonality Regularizer maintains the unitary properties of $U, Vt$. If the unitary properties are maintained, the angular alignment and any angular transformation will be meaningful (preserving lengths and angles). Hence, removing the regularizer negatively impacts both accuracy and computational efficiency, which might worsen in deeper models. Overall, the ablation study demonstrates that each loss component is essential for SSA's performance and efficiency.

\subsection{Comparison with DFA and its variants}
SSA introduces two key distinctions from DFA: the use of a structured weight-space and custom loss components applied directly in this SVD-space. To evaluate these differences, we present comparisons between SSA and DFA, along with its variants ~\cite{sanfiz2021benchmarking}, in the following tables and figures. The variants are explained in Section 3.

\begin{figure}
    \centering
    \includegraphics[width=0.75\linewidth]{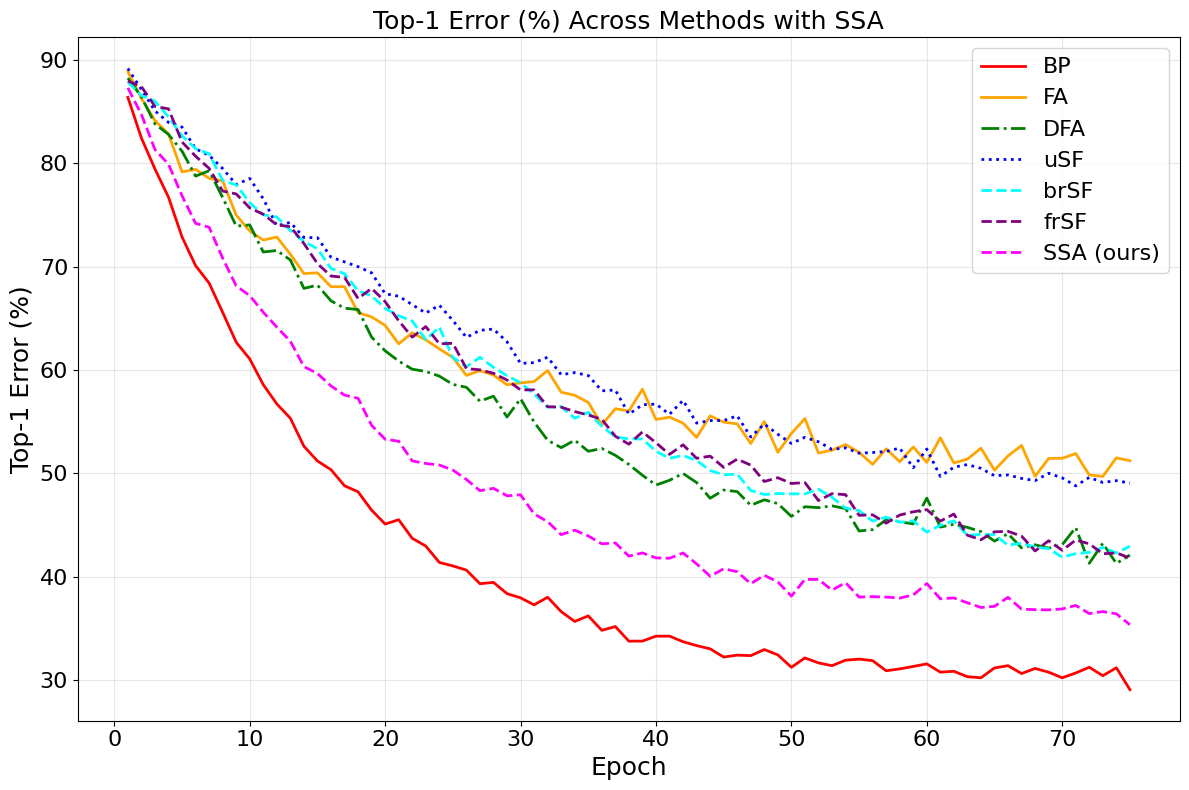} 
    \caption{\small Top-1 Error (\%) Across Epochs for a 3-layer MLP.}
    \label{fig:mlp_error}
    \vspace{-5mm}
\end{figure}

\begin{table}[htbp]
    \centering
    \small
    \setlength{\tabcolsep}{5pt} 
    \begin{tabular}{l|c|c|c|c}
        \hline
        \textbf{Method} & \textbf{LeNet} & \textbf{ResNet-20} & \textbf{ResNet-56} & \textbf{ResNet-18 (I)} \\ \hline
        BP     & 15.92 & 10.01 & 7.83 & 30.39 \\
        FA     & 40.67 & 29.59 & 29.23 & 85.25 \\
        DFA    & 37.59 & 32.16 & 32.02 & 82.45 \\
        uSF    & 16.34 & 10.59 & 9.19 & 34.97 \\
        brSF   & 17.08 & 11.08 & 10.13 & 37.21 \\
        SSA (Ours) & 16.27 & 10.63 & 9.72 & 33.41 \\ \hline
    \end{tabular}
    \caption{\small CIFAR-10 test error (\%) for LeNet, ResNet-20, and ResNet-56, and ImageNet Top-1 error (\%) for ResNet-18 (I). SSA consistently outperforms DFA and FA, narrowing the gap to BP.}
    \label{tab:combined_results}
\end{table}
We evaluate SSA and DFA (with its variants) on CIFAR-10 and ImageNet datasets. Tables \ref{tab:combined_results} summarize the test error rates for these methods, including the baseline BP. Our results show that SSA outperforms most variants of DFA across both datasets. To further analyze convergence behavior, we plot the error across epochs for a 3-layer MLP trained with SSA, BP, and DFA (with its variants) in Figure \ref{fig:mlp_error}. The results illustrate that SSA converges significantly faster than DFA and its variants, showcasing the advantages of structured feedback and custom loss design. While DFA and its variants perform poorly on convolutional layers, or cannot be directly applied to them, SSA achieves robust performance across both fully connected and convolutional architectures. To ensure uniformity in comparisons, the plotted results focus on MLPs. This limitation of DFA on convolutional layers further highlights the versatility of SSA for broader network types. Note, we also achieve alignment angles between SSA and BP gradients better than that of DFA gradients, as seen in Section 5.4.

\subsection{Memory and Compute Cost Analysis}
SSA performs inference using the decomposed form \( W_i = U_i S_i V_i^\top \), avoiding reconstruction of full weight matrices. The forward computation involves:
$\mathcal{O}(r \cdot n)$ for $V^\top x$, \;
$\mathcal{O}(r)$ for scaling by $S$, \;
$\mathcal{O}(m \cdot r)$ for projection by $U$,
leading to an overall inference complexity of \( \mathcal{O}(r(n + m + 1)) \), where \( r \ll \min(m, n) \). During inference, SSA stores only the compact low-rank factors:
$\mathcal{O}(m \cdot r) + \mathcal{O}(r) + \mathcal{O}(r \cdot n)$,
which is significantly smaller than storing the full dense weight matrix of size \( \mathcal{O}(m \cdot n) \).  
The SVD decomposition itself is performed only once prior to training, with complexity \(\mathcal{O}(m n \min(m, n))\), and this cost is negligible relative to the total training time dominated by repeated forward and backward passes.

\begin{figure}[h]
\centering
\includegraphics[width=0.7\columnwidth]{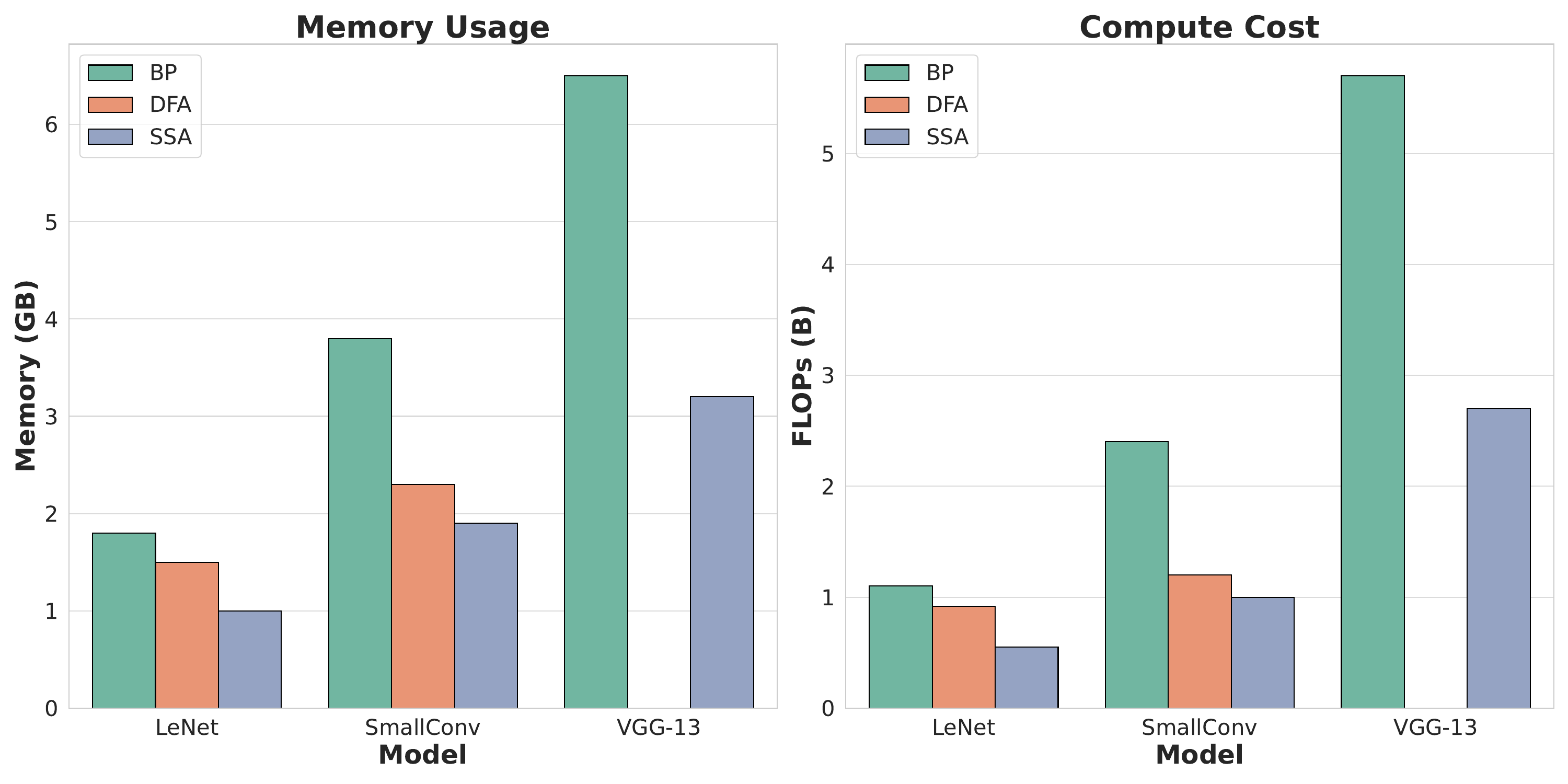}
\vspace{-3mm}
\caption{Compute and memory cost comparison of BP, DFA, and SSA across different architectures. SSA achieves lowest inference-time cost through structured low-rank training.}
\label{fig:dfa_mem}
\vspace{-3mm}
\end{figure}

\textbf{Empirical Comparison.}
Figure~\ref{fig:dfa_mem} compares compute and memory costs of SSA, DFA, and BP on LeNet, SmallConv, and VGG-13. DFA is efficient on shallow models but fails to scale, whereas SSA sustains efficiency on deeper networks and achieves lower inference-time cost through structure-aware low-rank training.

\section{Conclusion}
We introduced \textit{SVD-Space Alignment} (SSA), a novel local learning framework that combines Singular Value Decomposition (SVD) with Direct Feedback Alignment (DFA) to enable efficient, scalable training of deep neural networks. Unlike prior DFA-based methods, which failed to scale to convolutional or deep architectures, SSA incorporates structure into both the forward and feedback pathways, making local learning viable even for models like ResNet. By performing updates directly in the SVD-space, SSA utilizes a structured local loss function that ensures gradient alignment, stability through orthogonality preservation, and efficient convergence. The projection of gradients onto the Stiefel manifold complements the orthogonality regularizer, maintaining the geometry of the SVD-space while enhancing training robustness. Our theoretical analysis demonstrated that SSA gradients align positively with backpropagation gradients, guaranteeing descent directions, while dynamic rank reduction enables lightweight inference in low-rank subspaces. Experiments show that SSA achieves competitive accuracy on par with backpropagation, with substantial reductions in memory and compute. Ablation studies underscore the importance of each loss component in balancing accuracy and efficiency. In summary, SSA offers a principled and scalable local learning method, paving the way for resource-aware neural network training in real-world applications.

{
    \small
    \bibliographystyle{ieeenat_fullname}
    \bibliography{main}

\begin{thebibliography}{29}
\providecommand{\natexlab}[1]{#1}
\providecommand{\url}[1]{\texttt{#1}}
\expandafter\ifx\csname urlstyle\endcsname\relax
  \providecommand{\doi}[1]{doi: #1}\else
  \providecommand{\doi}{doi: \begingroup \urlstyle{rm}\Url}\fi

\bibitem[Ablin and Peyr{\'e}(2022)]{ablin2022fast}
Pierre Ablin and Gabriel Peyr{\'e}.
\newblock Fast and accurate optimization on the orthogonal manifold without retraction.
\newblock In \emph{International Conference on Artificial Intelligence and Statistics}, pages 5636--5657. PMLR, 2022.

\bibitem[Akrout et~al.(2019)Akrout, Wilson, Humphreys, Lillicrap, and Tweed]{akrout2019deep}
Mohamed Akrout, Collin Wilson, Peter Humphreys, Timothy Lillicrap, and Douglas~B Tweed.
\newblock Deep learning without weight transport.
\newblock \emph{Advances in neural information processing systems}, 32, 2019.

\bibitem[Belilovsky et~al.(2019)Belilovsky, Eickenberg, and Oyallon]{belilovsky2019greedy}
Eugene Belilovsky, Michael Eickenberg, and Edouard Oyallon.
\newblock Greedy layerwise learning can scale to imagenet.
\newblock In \emph{International conference on machine learning}, pages 583--593. PMLR, 2019.

\bibitem[Bordelon and Pehlevan(2022)]{bordelon2022influence}
Blake Bordelon and Cengiz Pehlevan.
\newblock The influence of learning rule on representation dynamics in wide neural networks.
\newblock \emph{arXiv preprint arXiv:2210.02157}, 2022.

\bibitem[Boyd and Vandenberghe(2004)]{boyd2004convex}
Stephen Boyd and Lieven Vandenberghe.
\newblock \emph{Convex Optimization}.
\newblock Cambridge University Press, 2004.

\bibitem[Crafton et~al.(2019)Crafton, Parihar, Gebhardt, and Raychowdhury]{crafton2019direct}
Brian Crafton, Abhinav Parihar, Evan Gebhardt, and Arijit Raychowdhury.
\newblock Direct feedback alignment with sparse connections for local learning.
\newblock \emph{Frontiers in neuroscience}, 13:\penalty0 525, 2019.

\bibitem[Dellaferrera and Kreiman(2022)]{dellaferrera2022error}
Giorgia Dellaferrera and Gabriel Kreiman.
\newblock Error-driven input modulation: Solving the credit assignment problem without a backward pass.
\newblock In \emph{International Conference on Machine Learning}, pages 4937--4955. PMLR, 2022.

\bibitem[Denil et~al.(2013)Denil, Shakibi, Dinh, Ranzato, and De~Freitas]{denil2013predicting}
Misha Denil, Babak Shakibi, Laurent Dinh, Marc'Aurelio Ranzato, and Nando De~Freitas.
\newblock Predicting parameters in deep learning.
\newblock \emph{Advances in neural information processing systems}, 26, 2013.

\bibitem[Denton et~al.(2014)Denton, Zaremba, Bruna, LeCun, and Fergus]{denton2014exploiting}
Emily~L Denton, Wojciech Zaremba, Joan Bruna, Yann LeCun, and Rob Fergus.
\newblock Exploiting linear structure within convolutional networks for efficient evaluation.
\newblock \emph{Advances in neural information processing systems}, 27, 2014.

\bibitem[Frenkel et~al.(2021)Frenkel, Lefebvre, and Bol]{frenkel2021learning}
Charlotte Frenkel, Martin Lefebvre, and David Bol.
\newblock Learning without feedback: Fixed random learning signals allow for feedforward training of deep neural networks.
\newblock \emph{Frontiers in neuroscience}, 15:\penalty0 629892, 2021.

\bibitem[He et~al.(2016)He, Zhang, Ren, and Sun]{he2016deep}
Kaiming He, Xiangyu Zhang, Shaoqing Ren, and Jian Sun.
\newblock Deep residual learning for image recognition.
\newblock In \emph{Proceedings of the IEEE conference on computer vision and pattern recognition}, pages 770--778, 2016.

\bibitem[Hinton(2022)]{hinton2022forward}
Geoffrey Hinton.
\newblock The forward-forward algorithm: Some preliminary investigations.
\newblock \emph{arXiv preprint arXiv:2212.13345}, 2022.

\bibitem[Kingma(2014)]{kingma2014adam}
Diederik~P Kingma.
\newblock Adam: A method for stochastic optimization.
\newblock \emph{arXiv preprint arXiv:1412.6980}, 2014.

\bibitem[Krizhevsky et~al.(2009)Krizhevsky, Hinton, et~al.]{krizhevsky2009learning}
Alex Krizhevsky, Geoffrey Hinton, et~al.
\newblock Learning multiple layers of features from tiny images.
\newblock 2009.

\bibitem[Krizhevsky et~al.(2017)Krizhevsky, Sutskever, and Hinton]{krizhevsky2017imagenet}
Alex Krizhevsky, Ilya Sutskever, and Geoffrey~E Hinton.
\newblock Imagenet classification with deep convolutional neural networks.
\newblock \emph{Communications of the ACM}, 60\penalty0 (6):\penalty0 84--90, 2017.

\bibitem[Launay et~al.(2020)Launay, Poli, Boniface, and Krzakala]{launay2020direct}
Julien Launay, Iacopo Poli, Fran{\c{c}}ois Boniface, and Florent Krzakala.
\newblock Direct feedback alignment scales to modern deep learning tasks and architectures.
\newblock \emph{Advances in neural information processing systems}, 33:\penalty0 9346--9360, 2020.

\bibitem[Lillicrap et~al.(2016)Lillicrap, Cownden, Tweed, and Akerman]{lillicrap2016random}
Timothy~P Lillicrap, Daniel Cownden, Douglas~B Tweed, and Colin~J Akerman.
\newblock Random synaptic feedback weights support error backpropagation for deep learning.
\newblock \emph{Nature communications}, 7\penalty0 (1):\penalty0 13276, 2016.

\bibitem[Lillicrap et~al.(2020)Lillicrap, Santoro, Marris, Akerman, and Hinton]{lillicrap2020backpropagation}
Timothy~P Lillicrap, Adam Santoro, Luke Marris, Colin~J Akerman, and Geoffrey Hinton.
\newblock Backpropagation and the brain.
\newblock \emph{Nature Reviews Neuroscience}, 21\penalty0 (6):\penalty0 335--346, 2020.

\bibitem[Ma et~al.(2024)Ma, Wu, Si, and Tan]{ma2024scaling}
Chenxiang Ma, Jibin Wu, Chenyang Si, and Kay~Chen Tan.
\newblock Scaling supervised local learning with augmented auxiliary networks.
\newblock \emph{arXiv preprint arXiv:2402.17318}, 2024.

\bibitem[Neftci et~al.(2017)Neftci, Augustine, Paul, and Detorakis]{neftci2017event}
Emre~O Neftci, Charles Augustine, Somnath Paul, and Georgios Detorakis.
\newblock Event-driven random back-propagation: Enabling neuromorphic deep learning machines.
\newblock \emph{Frontiers in neuroscience}, 11:\penalty0 324, 2017.

\bibitem[N{\o}kland(2016)]{nokland2016direct}
Arild N{\o}kland.
\newblock Direct feedback alignment provides learning in deep neural networks.
\newblock \emph{Advances in neural information processing systems}, 29, 2016.

\bibitem[N{\o}kland and Eidnes(2019)]{nokland2019training}
Arild N{\o}kland and Lars~Hiller Eidnes.
\newblock Training neural networks with local error signals.
\newblock In \emph{International conference on machine learning}, pages 4839--4850. PMLR, 2019.

\bibitem[Refinetti et~al.(2021)Refinetti, d’Ascoli, Ohana, and Goldt]{refinetti2021align}
Maria Refinetti, St{\'e}phane d’Ascoli, Ruben Ohana, and Sebastian Goldt.
\newblock Align, then memorise: the dynamics of learning with feedback alignment.
\newblock In \emph{International Conference on Machine Learning}, pages 8925--8935. PMLR, 2021.

\bibitem[Sanfiz and Akrout(2021)]{sanfiz2021benchmarking}
Albert~Jim{\'e}nez Sanfiz and Mohamed Akrout.
\newblock Benchmarking the accuracy and robustness of feedback alignment algorithms.
\newblock \emph{arXiv preprint arXiv:2108.13446}, 2021.

\bibitem[Schotth{\"o}fer et~al.(2022)Schotth{\"o}fer, Zangrando, Kusch, Ceruti, and Tudisco]{schotthofer2022low}
Steffen Schotth{\"o}fer, Emanuele Zangrando, Jonas Kusch, Gianluca Ceruti, and Francesco Tudisco.
\newblock Low-rank lottery tickets: finding efficient low-rank neural networks via matrix differential equations.
\newblock \emph{Advances in Neural Information Processing Systems}, 35:\penalty0 20051--20063, 2022.

\bibitem[Shalev-Shwartz and Ben-David(2014)]{shalev2014understanding}
Shai Shalev-Shwartz and Shai Ben-David.
\newblock \emph{Understanding Machine Learning: From Theory to Algorithms}.
\newblock Cambridge University Press, 2014.

\bibitem[Simonyan and Zisserman(2014)]{simonyan2014very}
Karen Simonyan and Andrew Zisserman.
\newblock Very deep convolutional networks for large-scale image recognition.
\newblock \emph{arXiv preprint arXiv:1409.1556}, 2014.

\bibitem[Webster et~al.(2021)Webster, Choi, et~al.]{webster2021learning}
Matthew~Bailey Webster, Jonghyun Choi, et~al.
\newblock Learning the connections in direct feedback alignment.
\newblock 2021.

\bibitem[Yang et~al.(2020)Yang, Tang, Wen, Yan, Hu, Li, Li, and Chen]{yang2020learning}
Huanrui Yang, Minxue Tang, Wei Wen, Feng Yan, Daniel Hu, Ang Li, Hai Li, and Yiran Chen.
\newblock Learning low-rank deep neural networks via singular vector orthogonality regularization and singular value sparsification.
\newblock In \emph{Proceedings of the IEEE/CVF conference on computer vision and pattern recognition workshops}, pages 678--679, 2020.

\end{thebibliography}
}

\end{document}